\documentclass[journal]{IEEEtran}
\usepackage{blindtext}
\usepackage{amsmath}
\usepackage{subcaption}
\usepackage{caption}
\usepackage[export]{adjustbox}
\usepackage{graphicx}
\usepackage{diagbox}
\usepackage[table,xcdraw]{xcolor}
\usepackage{url}

\usepackage[]{microtype} 

\usepackage{cite} 

\usepackage[normalem]{ulem}
\useunder{\uline}{\ul}{}

\usepackage{framed,multirow}

\ifCLASSINFOpdf
\else
\fi
\newcommand{\rr}{\mathop{\mathrm{RR}}}

\begin{document}

%
\title{
Feature point detection in HDR images based on coefficient of variation
}
%
%
%
\author{

\IEEEauthorblockN{Artur Santos Nascimento*, Welerson Augusto Lino de Jesus Melo, \\ Daniel Oliveira Dantas, Beatriz Trinchão Andrade}\\
\IEEEauthorblockA{Department of Computing, Federal University of Sergipe 
(UFS), São Cristóvão, Brazil\\
\{artur.nascimento, welerson.melo, ddantas, beatriz\}@dcomp.ufs.br}
%
%
}
\maketitle

\begin{abstract}
Feature point (FP) detection is a fundamental step of many computer vision tasks. However, FP detectors are usually designed for low dynamic range (LDR) images. In scenes with extreme light conditions, LDR images present saturated pixels, which degrade FP detection. On the other hand, high dynamic range (HDR) images usually present no saturated pixels but FP detection algorithms do not take advantage of all the information present in such images. FP detection frequently relies on differential methods, which work well in LDR images. However, in HDR images, the differential operation response in bright areas overshadows the response in dark areas. As an alternative to standard FP detection methods, this study proposes an FP detector based on a coefficient of variation (CV) designed for HDR images. The CV operation adapts its response based on the standard deviation of pixels inside a window, working well in both dark and bright areas of HDR images. The proposed and standard detectors are evaluated by measuring their repeatability rate (RR) and uniformity. Our proposed detector shows better performance when compared to other standard state-of-the-art detectors. In uniformity metric, our proposed detector surpasses all the other algorithms. In other hand, when using the repeatability rate metric, the proposed detector is worse than Harris for HDR and SURF detectors.


\end{abstract}


%
\IEEEpeerreviewmaketitle


\section{Introduction}
\label{sec:introduction}

Some computer vision tasks, such as image stitching, object tracking, face recognition and 3D reconstruction, rely on detection, description, and matching of feature points (FPs)~\cite{schmid2000, szeliski2011computerVision, liu2015uniting, oussalah2014hdr, korshunov2015impact, andrade2012}. FP detection algorithms are designed for low dynamic range (LDR) images. LDR images are represented using gamma-corrected 8-bit integer representation. Due to this characteristic, the performance of FP detectors decreases in LDR images of scenes with high lighting variations~\cite{zhou2016evaluating}. In scenes with extreme light conditions, LDR images become too bright in overexposed areas or too dark in underexposed areas, losing scene information. To overcome these problems, high dynamic range (HDR) images can be used. In HDR images, light acquired is proportional to the physical luminance of the scene, preserving details in dark and bright areas, which can potentially improve FP detection~\cite{rana2015evaluation}.

HDR images have been successfully applied to several computational problems: video encoding~\cite{mai2013visually}, inverse tone mapping~\cite{wang2015pseudo}, quality assessment~\cite{hadizadeh2017full}, and detection of salience~\cite{dong2016human}. An HDR image can be converted to an LDR image through a process known as tone-mapping. 
Tone-mapping is a process that creates an LDR image compressed from an HDR image and aims to preserve the best characteristics of bright and dark areas of the scene~\cite{mantiuk2015high}.

Assuming that detectors are highly dependent on scene illumination and light variation~\cite{mikolajczyk2005afine}, HDR and Tone-mapped (TM) images have the potential to be more robust and achieve better results in FP detection with extreme light conditions compared to LDR images. In this paper, we study how HDR-based methods can be used to improve FP detection in scenes with extreme light conditions.

P{\v{r}}ibyl et al.~\cite{pvribyl2016,pvribyl2013} evaluated how well FPs are repeated and distributed in LDR, TM, and HDR images from the same scene. Rana et al.~\cite{rana2016optimizing} proposed a new approach to improve detection in TM images and compared the results with LDR and HDR images. Kontogianni et al.~\cite{kontogianni2015hdr} applied HDR to improve detection in scenes with extreme light conditions. All these studies have shown that detecting FP directly in HDR images did not provide better results as expected compared with LDR images, and in most cases, they provided poorer results compared with TM images. Some studies have shown experimentally that
HDR significantly biases the localization of FP towards extreme bright areas, being one of the causes of the poor results mentioned above~\cite{rana2015evaluation, rana2016optimizing, deMelo2018improving}.
Other recent studies~\cite{suma2016evaluation, rana2016optimizing, rana2018learning} have focused on improving detection by modifying the tone-mapping operator (TMO). 

Our previous study~\cite{deMelo2018improving} has shown that standard FP detection algorithms (Harris Corner~\cite{harris1988combined} and DoG~\cite{lowe2004distinctive}) can be modified to improve detection in HDR images by using the coefficient of variation mask (CVM). CVM response is similar to derivative operations traditionally used in FP detectors and has the advantage that the response does not increase with pixel intensity, depending only on the variation of pixel intensities inside a window. Results have shown that FP detection in HDR images can perform better than in both LDR images and most TM images.

P{\v{r}}ibyl et al.~\cite{pvribyl2016, pvribyl2013} evaluated how well FPs are repeated and distributed in LDR, TM, and HDR images from the same scene. Rana et al.~\cite{rana2016optimizing} proposed a new approach to improve detection in TM images and compared the results with LDR and HDR images. Kontogianni et al.~\cite{kontogianni2015hdr} applied HDR to improve detection in scenes with extreme light selected the one with the best results.

In this study, we propose a new detector algorithm, called DetectorCV, designed specially to detect FPs in HDR images. To the best of our knowledge, no work in the literature has proposed a detector algorithm designed from scratch to deal with the problem of FP detection in HDR images. We tested different combinations of CVM with image filters and transformations as steps of the algorithm and selected the one with the best result. We then applied the optimized algorithm and the traditional ones to HDR images with extreme light conditions. Performance stability and distribution of the FP detected was measured by calculating repeatability Rate~\cite{schmid2000}, Uniformity~\cite{deMelo2018improving}, and vector dominance~\cite{coello2007evolutionary}. Results showed that usage of CVM as a base of an FP detection algorithm for HDR image improves results in comparison with both traditional and modified algorithms proposed in~\cite{deMelo2018improving}

\section{Background}
\label{sec:background}

In this section, we show how detector algorithms used in the experiments work. We also define terms related to HDR imagery as well as the coefficient of variation mask (CVM).

\subsection{Detector algorithms}

FP detection algorithms are an initial step in the image feature extraction pipeline. Detection algorithms are designed to find relevant points, which can be repeatedly detected in different images, even when some geometric or projective transformation has occurred. An FP detector is often combined with an FP descriptor algorithm, which extracts a local descriptor, i.e., a vector of characteristics, for each detected point. These vectors are used to match points from different images. See~\cite{mikolajczyk2005performance} for a review of the descriptor algorithms.

FP detection algorithms can be divided into two categories: corner detectors and blob detectors~\cite{mikolajczyk2005performance}. In this paper, we consider three popular algorithms: Harris Corner Detector~\cite{harris1988combined}; Difference of Gaussian (DoG), which is the detector of the scale-invariant feature transform (SIFT) algorithm~\cite{lowe2004distinctive}; and the detector of speeded-up robust features (SURF) feature extractor algorithm~\cite{bay2008speeded}. Harris Corner and DoG can be categorized as a corner detector, and SURF as a blob detector.

In a previous study~\cite{deMelo2018improving}, we proposed adaptations in the Harris Corner and DoG algorithm to improve detection in HDR images. These algorithms are called Harris Corner for HDR, and DoG for HDR. Following the same methodology applied to the DoG for HDR algorithm, we implemented the SURF for HDR in this current study. The six algorithms are summarized as follows:

\begin{itemize}
    \item \textbf{Harris Corner}~\cite{harris1988combined}. Harris Corner detector is based on the local auto-correlation function reflecting local intensity changes in the image. The algorithm has three main steps. (1) \textit{Gaussian filter}: convolve input image with a Gaussian filter. (2) \textit{Sobel operator}: obtain derivatives by applying Sobel operator on x-axis and y-axis and multiply its results to obtain the elements of a $2\times2$ Hessian matrix. (3) \textit{Resulting image}: an image is calculated from the Hessian matrix, a threshold is applied to obtain the best regions, and \textit{local maxima suppression} selects the FPs from these regions.
    
    \item \textbf{Harris Corner for HDR}~\cite{deMelo2018improving}. Harris Corner for HDR algorithm adds a new step between the step (1) \textit{Gaussian Filter} and (2) \textit{Sobel operator} of Harris Corner. The new step is the application of a coefficient of variation mask (CVM).
    
    \item \textbf{DoG}~\cite{lowe2004distinctive}. 
    Difference of Gaussian (DoG) algorithm is based on the scale-space technique and an approximation of Laplacian of Gaussian (LoG). The algorithm has three main steps. (1) \textit{Construct scale-space}: input image is resized to be able to find FPs in different scales. (2) \textit{LoG approximation}: calculate Difference of Gaussian, which is a derivative approach. (3) \textit{Find and threshold FPs}: \textit{local maxima suppression} in a $3\times3\times3$ neighborhood is used to select possible FPs, and a threshold is used to select the final FPs.
    
    \item \textbf{DoG for HDR}~\cite{deMelo2018improving}. In DoG for HDR, the CVM step is applied to the resized images at the beginning of step (1) \textit{Construct scale space} of the DoG algorithm.
    
    \item \textbf{SURF}~\cite{bay2008speeded}. SURF detector algorithm aims to be a detector as efficient as DoG, but more robust and fast. SURF is based on the Hessian matrix approximation and has three main steps. (1) \textit{Construct scale-space}: a fast scale-space is created by using integral images. (2) \textit{Fast Hessian}: is the derivative step that uses a Hessian approximation. (3) \textit{Find and threshold FPs}: \textit{local maxima suppression} in a $3\times3\times3$ neighborhood is used to select possible FPs, and a threshold is used to select the final FPs.
    
    \item \textbf{SURF for HDR}. SURF for HDR, similarly to DoG for HDR, just applies CVM to the resized images at the beginning of step (1) \textit{Construct scale space} of SURF.
    
\end{itemize}

\begin{figure}[t]
  \centering
  \includegraphics[width=\linewidth]{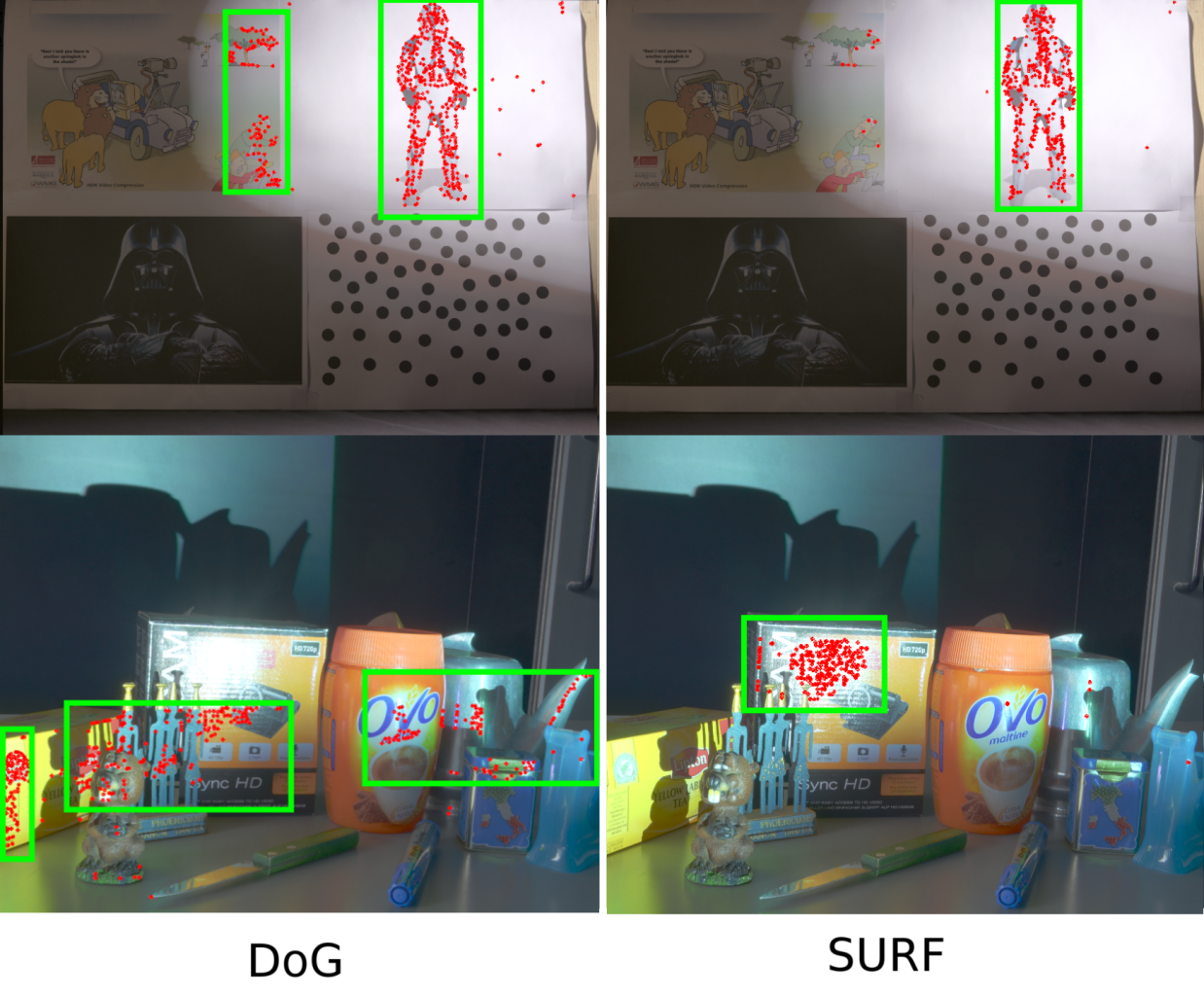}
  \caption{
  DoG and SURF detector applied to (first row) 2D Lighting~\cite{pvribyl2013}, and (second row) ProjectRoom~\cite{rana2015evaluation}  datasets. The FPs detected are marked with red dots in the scenes. The areas with a major concentration of FPs are highlighted with a green rectangle. In this figure, the HDR images were tone mapped using Mantiuk's operator~\cite{mantiuk2008display}.}
  \label{fig:ex_bad_keypoints_2D}
\end{figure}

\begin{figure*}[t]
    \centering
    \includegraphics[width=\linewidth]{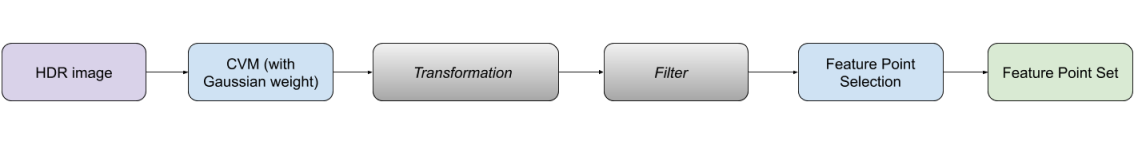}
    \caption{General flow chart of the DetectorCV algorithm.}
    \label{fig:detectorCV_flow}
\end{figure*}

\subsection{HDR imagery}


HDR images use floating-point values to represent pixels, allowing a larger dynamic range compared to LDR images~\cite{banterle}. HDR images can be generated through special devices able to capture more than eight bits per sample. However, it is more common to generate HDR images from a set of LDR images with different light exposure times.

The literature of HDR applied to computer vision problems is limited, and only recent studies have considered HDR or TM images as input data~\cite{rana2018learning}. Some recent studies include FP detection~\cite{rana2015evaluation, rana2016optimizing, pvribyl2016, pvribyl2013, deMelo2018improving, kontogianni2015hdr}, image matching~\cite{rana2018learning} and photogrammetric applications~\cite{ntregka2013photogrammetric}. Some works have also studied the impact of HDR content on privacy protection~\cite{vrevrabek2014evaluation}. Kontogianni~\cite{kontogianni2015hdr} studied HDR images of architectural scenes and compared them with LDR images. Agrafiotis~\cite{agrafiotis2015hdr} concludes that HDR images enhance people detection and tracking in indoor scenes. We can notice that the majority of these studies actually use TM images. The common conclusion is that TM images provide better results than HDR images, and HDR images provide better results than LDR images.

\subsection{Coefficient of variation mask}~\label{subsec:CVM}

The coefficient of variation (CV) of a population is the standard deviation divided by the average of the population~\cite{snedecor1967}. In the image processing field, the coefficient of variation mask (CVM) is a filter applied by calculating the CV of the neighborhood of each pixel. The CVM is used as a filter for some kinds of noises, as well as to find edges in images~\cite{yu2004, mora2005}. Consider as population a set $p$ of $n\times n$ pixels inside a window centered in pixel $(x, y)$.
The CV of $p$ is given by Equation~\ref{eq:cv}

\begin{equation}
 \operatorname{CV}(p) = \frac{\operatorname{\sigma}}{\mu}, 
 \label{eq:cv}
\end{equation}
\noindent
where $\mu$ is the arithmetic mean of the population, and $\operatorname{\sigma}$ is the standard deviation. The $\operatorname{\sigma}$ is defined in Equation~\ref{eq:sd}:

\begin{equation}
\operatorname{\sigma} = \sqrt{\frac{1}{N} \times\sum^N_{i=1}{(p_i - \mu)^2}},
\label{eq:sd}
\end{equation}
\noindent
where $N$ is the number of pixels inside the image mask, i.e., population size; and $p_i$ the i-th pixel of the mask.

Filters based on CV allow balanced and well-located edge strength measurement in both bright and dark regions~\cite{yu2004}, which makes CV suitable for the problem of detecting good FP in bright and dark regions of HDR images. Also, for different data sources or populations, the arithmetic mean and the standard deviation tend to change together; so, the CV is relatively stable or constant~\cite{snedecor1967}. This means that the resulting CV will be minimally affected by different intensities of light in the same image, since CV is a local operator.

Most detector algorithms, including Harris Corner, DoG and SURF, have a fundamental step based on some derivative of pixel values. Those algorithms were designed to assume a display-referred LDR image, usually gamma-corrected. In this case, the magnitude of the pixel's derivatives in dark and bright areas would not be significantly different. However, in the HDR image, where pixel values encode the scene's luminance, derivative responses increase significantly in bright areas. The derivative response in bright areas can be orders of magnitude greater than in dark areas~\cite{pvribyl2016}. Since detectors choose the FPs based on the highest derivative responses, most FPs are detected in extremely bright areas, leaving middle tones and dark areas with few or no FPs detected. As past studies have confirmed,~\cite{pvribyl2016, pvribyl2013, rana2015evaluation, deMelo2018improving}, most of the contemporary FP detector algorithms would process HDR inefficiently. 
Figure~\ref{fig:ex_bad_keypoints_2D} shows some examples of the problem mentioned above. We can observe that most of the $500$ FPs with highest responses chosen are in the bright area. Areas with highest concentration of FPs are highlighted with a green rectangle.

Applying CVM as a step in standard detector algorithms solves the problem of FP concentration in bright regions~\cite{deMelo2018improving}. However, this approach brings instability to the FPs detected, i.e., the repeatability rate metric was worse using CVM in some scenarios~\cite{deMelo2018improving}. The instability is caused by the derivative steps when executed after CVM, as the combination of CVM and a derivative step shifts the actual location of the FPs. 

\section{DetectorCV}
\label{sec:detectorCV}

DetectorCV is an FP detector algorithm based on CVM. The algorithm also involves some small adaptations in the CVM filter and a histogram transformation. In this section, we describe in detail the DetectorCV algorithm.

When using standard detector algorithms augmented with a CVM step, we noticed that FPs were usually displaced from the center of the mask. We also noticed that repeatability rate, one of the evaluation criteria, diminished due to this fact. To centralize the FPs and improve the detection, we applied a 2D Gaussian weight to the CVM. Instead of the standard deviation, defined by Equation~\eqref{eq:sd}, we used the variation $V$ defined by:

\begin{equation}
    V = \sqrt{\frac{1}{N} \times\sum^N_{i=1}{((p_i - \mu )^2)w_i}},
    \label{eq:sd2}
\end{equation}

\noindent
where $w_i$ is a weight used to place the detected feature point closer to the observed feature point in the image. The weight $w_i$ decreases with radius, having a maximum value in the center of the window. The weight $w_i$ is defined by:

\begin{equation}
w_i = G(x,y) = \frac{1}{2 \pi \sigma_c ^{2}} e^{- \frac{x^{2} + y^{2}}{2 \sigma_c ^{2}}},
 \label{eq:2dgauss}
\end{equation}

\noindent
where $(x,y)$ are coordinates inside the mask containing the pixels $p_i$ in Equation~\eqref{eq:sd2}.

\begin{figure}[ht!]
    \centering
    \begin{subfigure}[t]{\linewidth}
        \includegraphics[width=\linewidth]{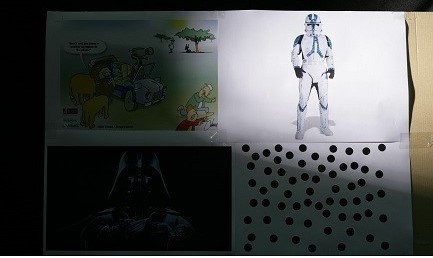}
        \caption{}
        \label{fig:sub:inputPribyl2D}
    \end{subfigure}
    
    \begin{subfigure}[t]{\linewidth}
        \includegraphics[width=\linewidth]{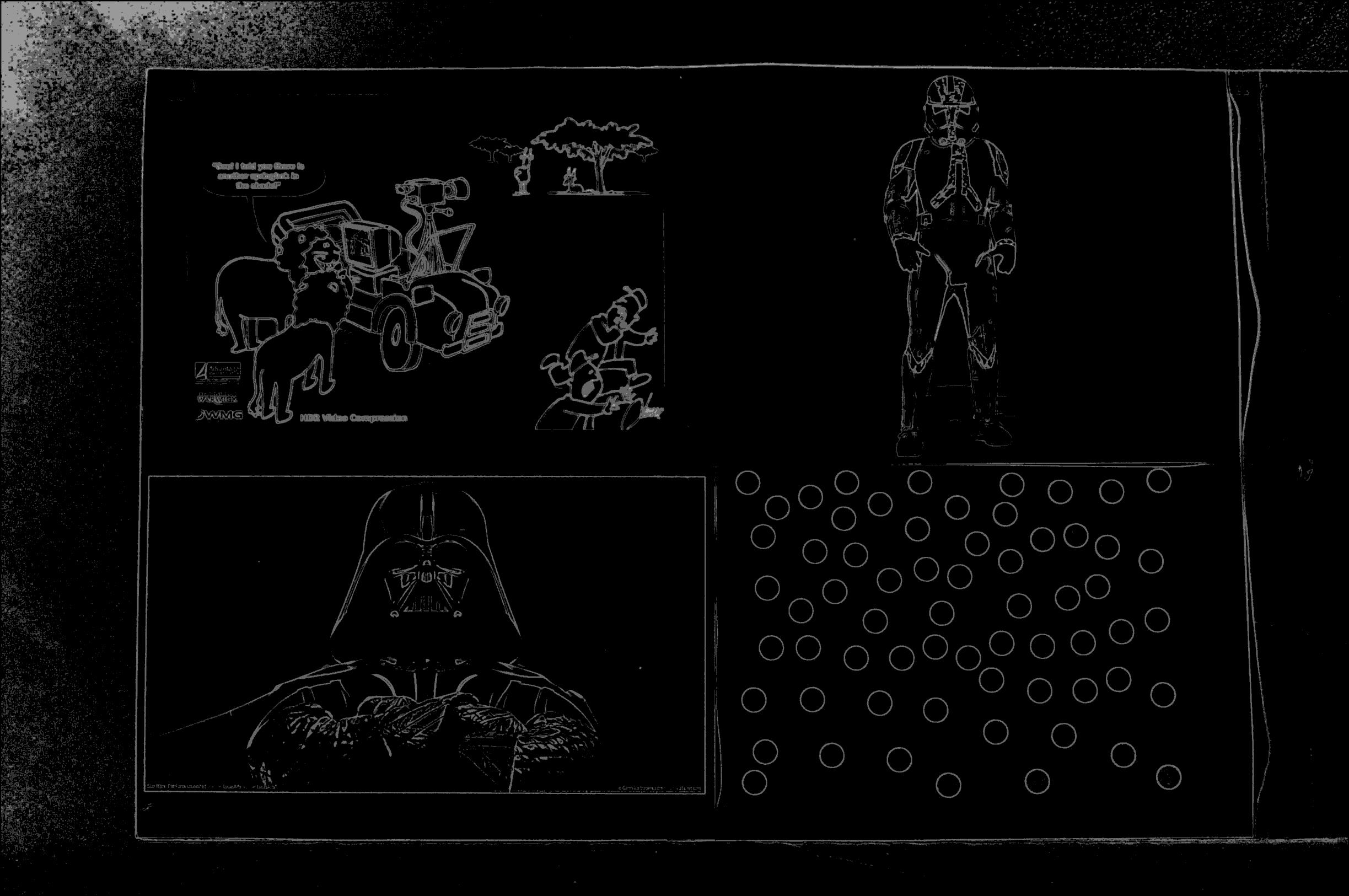}
        \caption{}
        \label{fig:sub:responseDetectorCV}
    \end{subfigure}
    
    \begin{subfigure}[t]{\linewidth}
        \includegraphics[width=\linewidth]{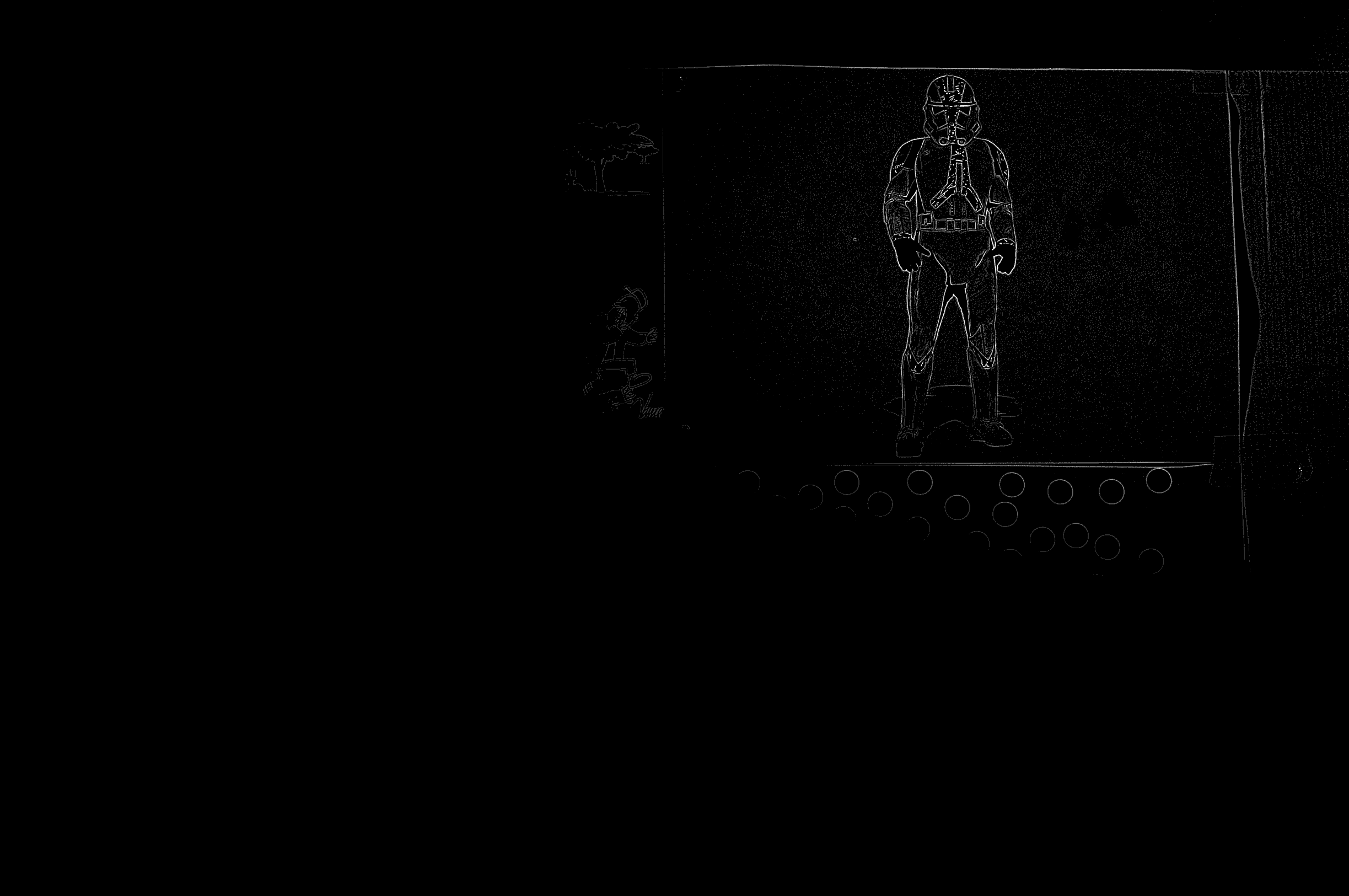}
        \caption{}
        \label{fig:sub:responseDoG}
    \end{subfigure}
    
    \caption{(a) Example of input image in 2D Lighting dataset of P{\v{r}}ibyl et al.~\cite{pvribyl2016}; (b) response of CVM with Gaussian weight; (c) Difference of Gaussian response}
    \label{fig:inputImagesPribyl}
\end{figure}

Figure~\ref{fig:sub:responseDetectorCV} shows an example of a resulting image of applying CVM with Gaussian weight.
It is noticeable that the scene's illumination does not affect the response image generated by the CVM.
Figure~\ref{fig:sub:responseDoG} shows an example of a resulting image of applying a DoG derivative step.
Here we can notice that the resulting image of the DoG detector loses details in dark regions of the image, as response is proportional to illumination. The number of detected FPs in dark regions is also very small, as can be seen in Figure~\ref{fig:ex_bad_keypoints_2D}.
Both examples have the Figure~\ref{fig:sub:inputPribyl2D} as input.

At first glance, CVM applied to the HDR images exhibited almost no noise. On the other hand, CVM applied to LDR images appeared much noisier. The CVM mask was used as an edge detector in LDR images because of the strong response for edges; it maintains similar behavior in HDR images. Because of this, points tended to be concentrated in regions with thick edges, which is not good for stability in detection, as it is poorly determined and therefore unstable to small amounts of noise~\cite{lowe2004distinctive}.
To deal with this problem, we applied:

\begin{itemize}
    \item a \textit{transformation} on the histogram, to simultaneously enhance contrast and highlight high contrast points, mainly corners; 
    \item a \textit{filter}, to reduce edge effect, preserve highlighted points, and reduce noise generated by histogram transformation.
\end{itemize} 

We chose three \textit{transformations} to test our approach. They are linear transformation, logarithmic transformation, and histogram equalization. For the next step, we chose two \textit{filters} to test out: Gaussian filter and bilateral filter. The definition of these transformations and filters can be found in the book by Gonzalez~\cite{gonzalez2008digital}.

Linear transformation was a single pixel-wise multiplication by a constant, defined empirically as $25$. Logarithmic transformation is represented by $c \times \mathrm{log}(f (x, y))$, where $f (x, y)$ is the input image and $c$ is an constant set, also empirically, to $150$. Histogram equalization is a well-known technique and does not have parameters. The empirical values mentioned above were reached by observing the results of a group of images from the datasets used.

After applying one of the transformations and one of the filters, we selected the FPs. The resulting image is submitted to a local maxima suppression step in a squared region to determine whether a pixel $(x, y)$ could be an FP. In the end, from all the FP candidates, the ones with the strongest responses are chosen as FPs.

We can summarize DetectorCV algorithm as follows: apply CVM with Gaussian weight; apply a \textit{transformation}; apply a \textit{filter}, and; apply feature point selection. Figure~\ref{fig:detectorCV_flow} shows the  general flow of the DetectorCV algorithm.

\section{Methodology}
\label{sec:method}

As seen in Figure~\ref{fig:detectorCV_flow}, the DetectorCV algorithm has four steps. After applying the CVM with Gaussian weight, there are \textit{Transformation} and \textit{Filter} steps. We used some well-known algorithms in the \textit{Transformation} and \textit{Filter} steps and evaluated different combinations to select the combination that gives the best result to our approach.


The main appeal of our approach is to detect points in areas with different intensities of illumination. We chose as input datasets with noticeable light variation, and we devised a criterion that evaluates the performance of an FP detector in such datasets. We compared all the possible combinations of the DetectorCV algorithm to select the combination that maximizes our FP evaluation criteria. This section details our experimental setup to perform this task.

\subsection{Dataset}
\label{sec:dataset}
\def\figw{4.2cm}
\begin{figure}[t]
    \centering
    \begin{subfigure}[t]{\figw}
        \includegraphics[width=\figw]{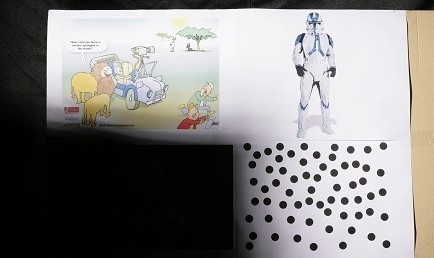}
        \caption{}
        \label{fig:sub:dataset2d}
    \end{subfigure}
    \begin{subfigure}[t]{\figw}
        \includegraphics[width=\figw]{datasets1b.jpg}
        \caption{}
        \label{fig:sub:dataset2d-b}
    \end{subfigure}
    
    \begin{subfigure}[t]{\figw}
        \includegraphics[width=\figw]{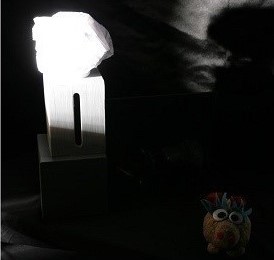}
        \caption{}
        \label{fig:sub:dataset3d}
    \end{subfigure}
    \begin{subfigure}[t]{\figw}
        \includegraphics[width=\figw]{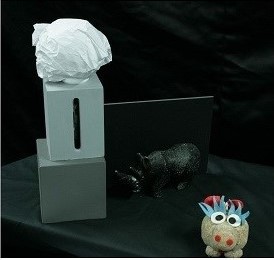}
        \caption{}
        \label{fig:sub:dataset3d-b}
    \end{subfigure}
    
    \begin{subfigure}[t]{\figw}
        \includegraphics[width=\figw]{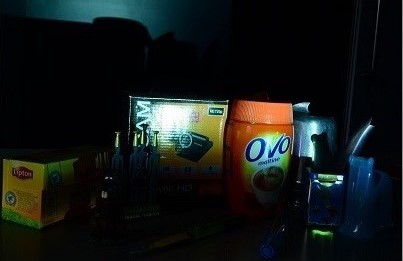}
        \caption{}
        \label{fig:sub:datasetRana}
    \end{subfigure}
    \begin{subfigure}[t]{\figw}
        \includegraphics[width=\figw]{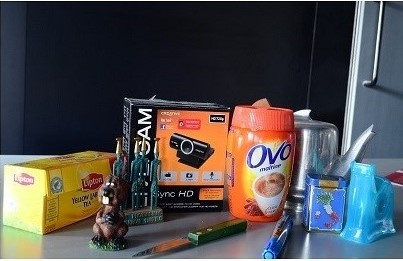}
        \caption{}
        \label{fig:sub:datasetRana-b}
    \end{subfigure}
    
    \caption{Examples of images from the datasets: (a) and (b) 2D Lighting (P{\v{r}}ibyl et al.~\cite{pvribyl2016}); (c) and (d) 3D Lighting (P{\v{r}}ibyl et al.~\cite{pvribyl2016}); and (e) and (f) ProjectRoom (Rana et al.~\cite{rana2016optimizing}). }
    \label{fig:imagesFromDatasets}
\end{figure}

In this paper, we used three image datasets: 2D Lighting (P{\v{r}}ibyl et al.~\cite{pvribyl2016}); 3D Lighting (P{\v{r}}ibyl et al.~\cite{pvribyl2016}); and ProjectRoom (Rana et al.~\cite{rana2016optimizing}). These datasets were created to evaluate the stability of detector algorithms when applied to pairs of images when suffering some change, such as distance, viewpoint or illumination. We used only HDR images with extreme illumination changes. 

The 2D Lighting dataset consists of images of a planar poster (Figure~\ref{fig:sub:dataset2d} and Figure~\ref{fig:sub:dataset2d-b}), while 3D Lighting dataset consists of images of objects with fine geometry and different colors to contrast together with the light (Figure~\ref{fig:sub:dataset3d} and Figure~\ref{fig:sub:dataset3d-b}). The 2D Lighting dataset contains images from scenes with distance, viewpoint, and illumination changes. The 3D Lighting dataset contains images from scenes with illumination changes. For each scene we have two types of image encoding: one being HDR, and the other being logHDR, which is an HDR image logarithmically encoded. The 2D Lighting dataset contains $36$ HDR images, and the 3D Lighting dataset contains $7$, which gives us a total of 43 HDR images. The number of logHDR images is also $43$.

Both 2D Lighting and 3D Lighting datasets were designed to contain three nonoverlapping areas with different illuminations. This fact allows us to evaluate the uniformity of FPs distribution among different illuminations in a single scene. The three areas were called, by P{\v{r}}ibyl et al.~\cite{pvribyl2016}, shadows, midtones, and highlights.

The ProjectRoom dataset contains seven HDR images from different scenes with extreme light variation. The scenes are composed of 3D objects placed on a table (Figure~\ref{fig:sub:datasetRana} and Figure~\ref{fig:sub:datasetRana-b}). The images in this dataset are not partitioned based on different illuminations. We propose an approach to partition the images in this database and to make our approach compatible with other datasets. The approach is described in Subsection~\ref{sec:image_partitioning}.

We used 2D Lighting and 3D Lighting datasets as input to define the best from the six possible combinations of DetectorCV algorithm, described in Section~\ref{sec:detectorCV}. ProjectRoom dataset is used later as input to compare the winner combination of the DetectorCV algorithm with detectors found in the literature.

\subsection{Evaluation Criteria}

The FP detector evaluation criteria chosen were two: repeatability rate (RR) proposed by Schimid~\cite{schmid2000}, and uniformity proposed in our past study~\cite{deMelo2018improving}.

\begin{itemize}
    \item \textbf{Repeatability rate} (RR), denoted by $\rr$ in Equation~\ref{eq:rr}, is a criterion in FP evaluation that is well known in literature. RR evaluates the stability of a detector algorithm among images of the same scene. Different images have some variation, such as distance to camera, light intensity, and viewpoint. RR resulting value is in interval $[0.0, 1.0]$, where the best result is $1.0$. A high RR value means that most FPs found in a reference image were found in a target image. 
    
    \begin{equation}
        \rr(I_r, I_t) = \frac{R_{rt}}{\min(n_r, n_t, M)},
        \label{eq:rr}
        \end{equation}
    \noindent
where $I_r$ is the reference image and $I_t$ is the test image. $n_r$ and $n_t$ are the total numbers of FPs detected in $I_r$ and $I_t$ respectively, $M$ is the maximum number of FPs to be considered, and $R_{rt}$ is the number of repeated FPs in both $I_r$ and $I_t$.

    \item \textbf{Uniformity}, represented as $U$ by Equation~\ref{eq:u}, evaluates how well distributed found FPs is among pre-divided areas in an image. We divided the image in areas based on scene illumination, so we can evaluate whether an FP detector has some bias for detecting FPs in dark or bright areas. Uniformity value is in interval $[0.0, 1.0]$, where the best result is $1.0$. The best result means that FPs found in the image are equally distributed among all pre-defined areas of the image, i.e., it is uniformly distributed. Uniformity is defined by:
    
    \begin{equation}
        U = 1-(\operatorname{maxFP} - \operatorname{minFP}) 
        \label{eq:u}
    \end{equation}
    
    \noindent
    where $\operatorname{maxFP}=\max(a_1/T,a_2/T,...,a_n/T)$ and $\operatorname{minFP}=\min(a_1/T,a_2/T,...,a_n/T)$. $T$ is the number of FPs detected, $n$ is the number of non-background areas in the image, $a_i$ is the number of FPs in the $i$-th area of the image, and $i$ is in range $[1, n]$.
        
\end{itemize}

With both criteria, we can evaluate how stable an FP detector algorithm is in relation to scene variations, and how stable the algorithm is in relation to the distribution of FPs among areas with different illuminations. A good detector must be able to have high values in both criteria.

\begin{figure}[t]
    \centering
    
    \begin{subfigure}[t]{0.98\linewidth}
        \includegraphics[width=\linewidth]{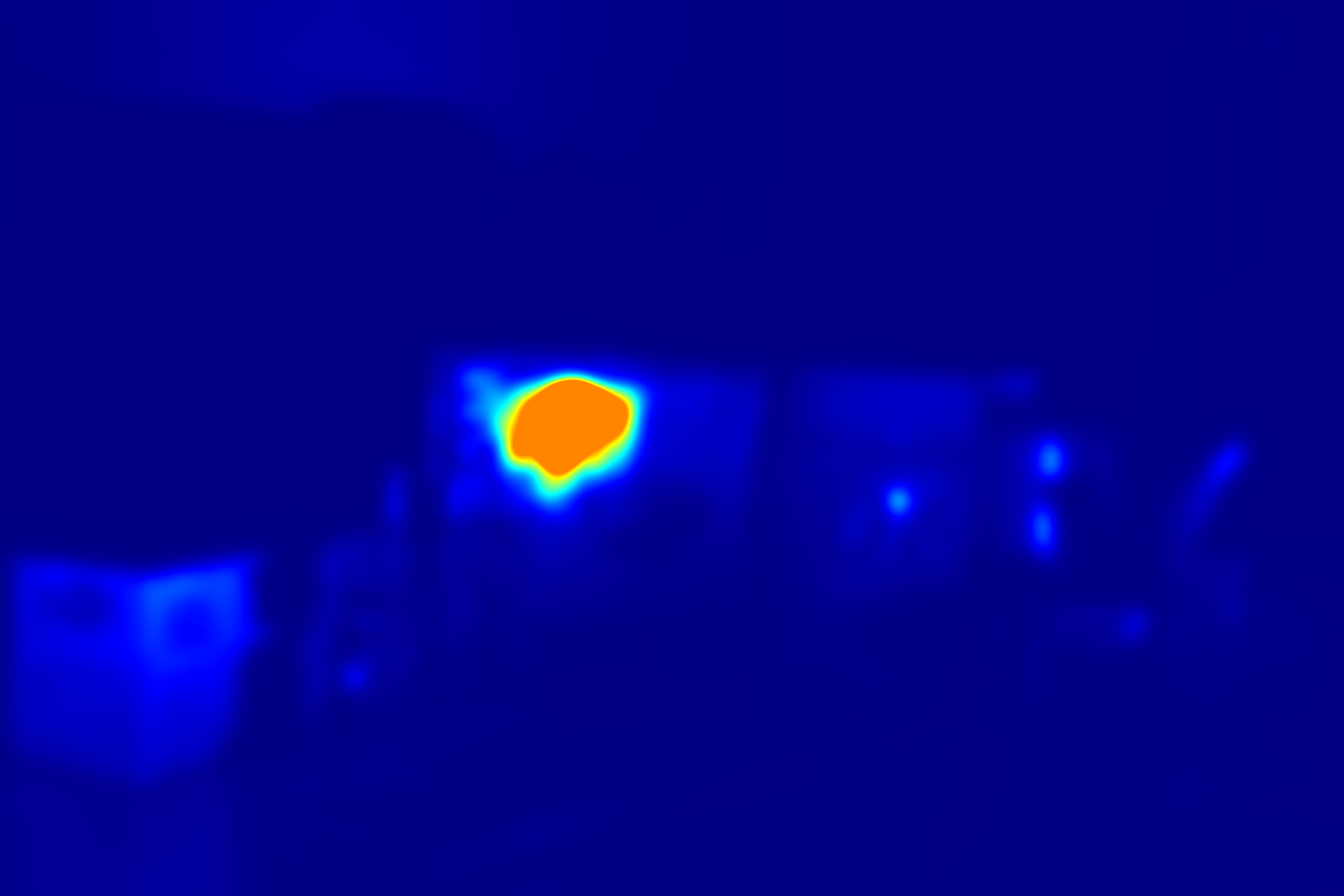}
        \caption{}
        \label{fig:luminanceMap}
    \end{subfigure}
    
    \begin{subfigure}[t]{0.98\linewidth}
        \includegraphics[width=\linewidth]{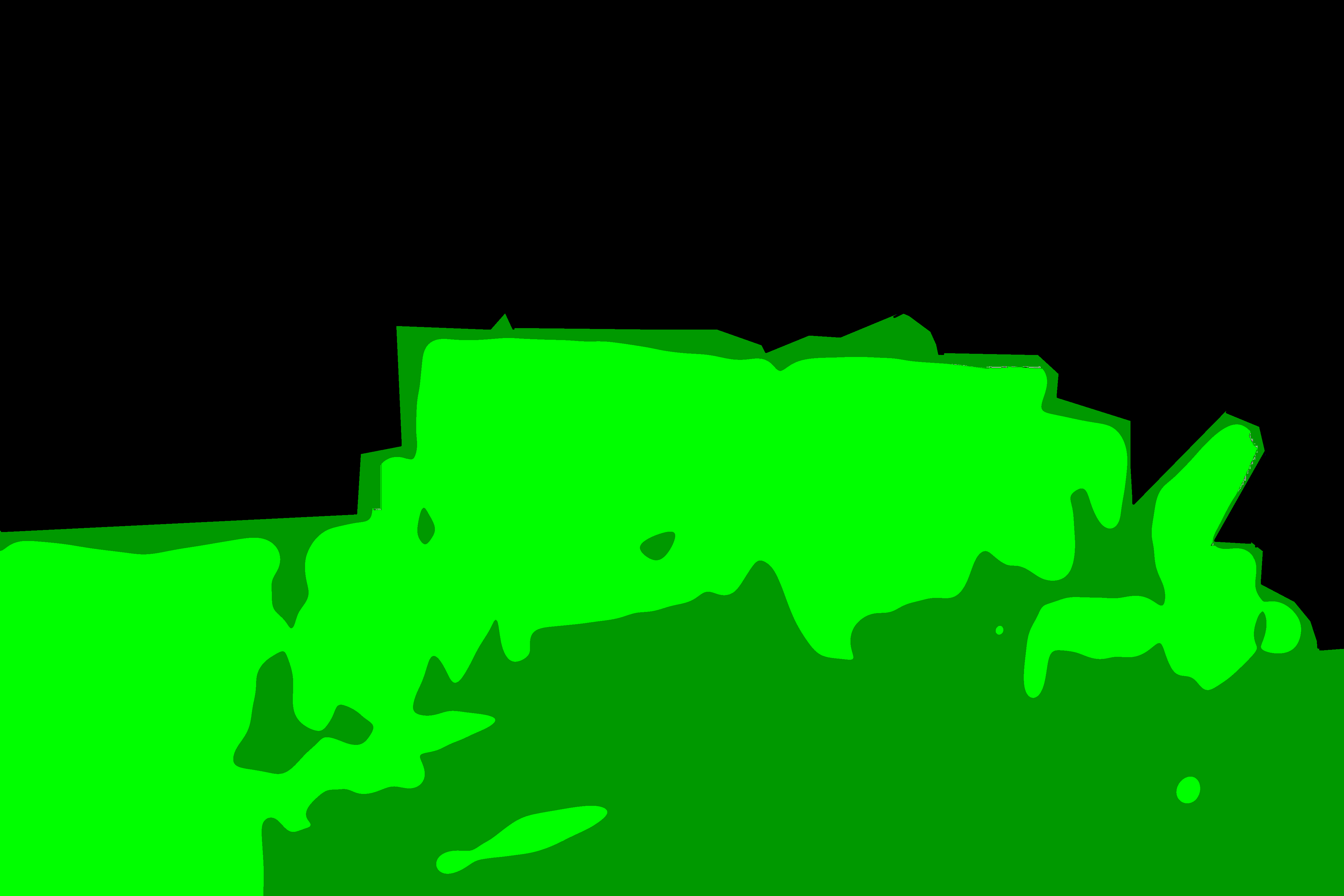}
        \caption{}
        \label{fig:divisionEx}
    \end{subfigure}
    
    \caption{Image partitioning in function of illumination: (a) luminance map image generated from Figure~\ref{fig:sub:datasetRana-b}; (b) final image divided: light green represents bright areas, dark green represents dark areas, and black represents the excluded background.}
\end{figure}

\subsubsection{Image partitioning in function of illumination} 
    \label{sec:image_partitioning}
    
The 2D Lighting dataset and 3D Lighting dataset scenes were designed to be easily divided into areas based on illumination levels. P{\v{r}}ibyl et al.~\cite{pvribyl2016} did it to apply an FP distribution evaluation technique. However, the division of image into areas based on illumination levels is not a common feature in datasets. ProjectRoom dataset did not have such division, i.e., it was not designed to have areas with constant levels of illumination. So, we needed to find a methodology to partition each image of the ProjectRoom dataset into different areas in function of illumination levels. As we could not find such a technique in the literature, we proposed one, as explained below.

The first step was to manually select the featureless areas in the image, i.e., the background. The background contains areas of the image without objects, therefore, usually not presenting detected FPs. These areas were ignored.

The remaining areas were then partitioned into two subareas. Each subarea, nonoverlapping and not necessarily connected, must have similar illumination characteristics: one with high luminance pixels and the other with low luminance pixels. A good technique to separate these subareas is fundamental to a good evaluation of uniformity. 
We used a luminance map to select $50\%$ of strongest pixels in the image as pixels with high luminance and the remaining ones as pixels with low luminance. This technique provided a good visual separation of objects where FPs could be found. It also resulted in two subareas with the same number of pixels, which allowed a fair uniformity evaluation.

In order to divide the image into subareas in function of scene illumination, we calculated a luminance map using the Retinex algorithm. In Retinex, luminance is estimated from the radiance map of the input image. As Retinex is a mathematically ill-posed problem~\cite{jobson1997properties}, it is not possible to obtain exact scene luminance. According to Retinex theory, an input image $I$ is a product of luminance $L$ of the scene (which varies with different illumination conditions) by the reflectance (which characterizes objects of the scene), i.e., $I = R L$~\cite{rana2016optimizing}.

In order to find L, we used the approach used by Chiu et al.~\cite{chiu1993spatially}. The image is defined as  $L = I * G_r$, where $G_r$ is a Gaussian filter with standard deviation calculated from the image size $[m\times n]$, i.e., $r = \alpha \cdot \max(m,n)$. The constant $\alpha$ was set to $0.007$, as in Rana et al.~\cite{rana2016optimizing}. Gaussian mask size was set as the smallest odd integer greater than $6r$. Afterwards, the image L was normalized to a 16-bit image.

In Figure~\ref{fig:luminanceMap} we can see an example of a luminance map rendered as a heat map. We calculated the cumulative histogram $H$ from the image $L$ ignoring the background pixels. We then defined the threshold $M$ to divide the feature rich area of the image in two subareas: \textit{bright area} and \textit{dark area}. A suitable $M$ will leave both subareas with approximately the same amount of pixels. Figure~\ref{fig:divisionEx} shows the image division in dark and bright areas from the luminance map of Figure~\ref{fig:luminanceMap}.

This solution is suitable for images with extreme light conditions, where it is expected that scenes will have some dark and bright areas. It is relevant to stress that this approach makes the uniformity evaluation criteria compatible with datasets that are not previously partitioned based on scene illumination.

\subsection{Selection test}

\begin{figure*}[ht!]
    \centering
    \includegraphics[width=\linewidth]{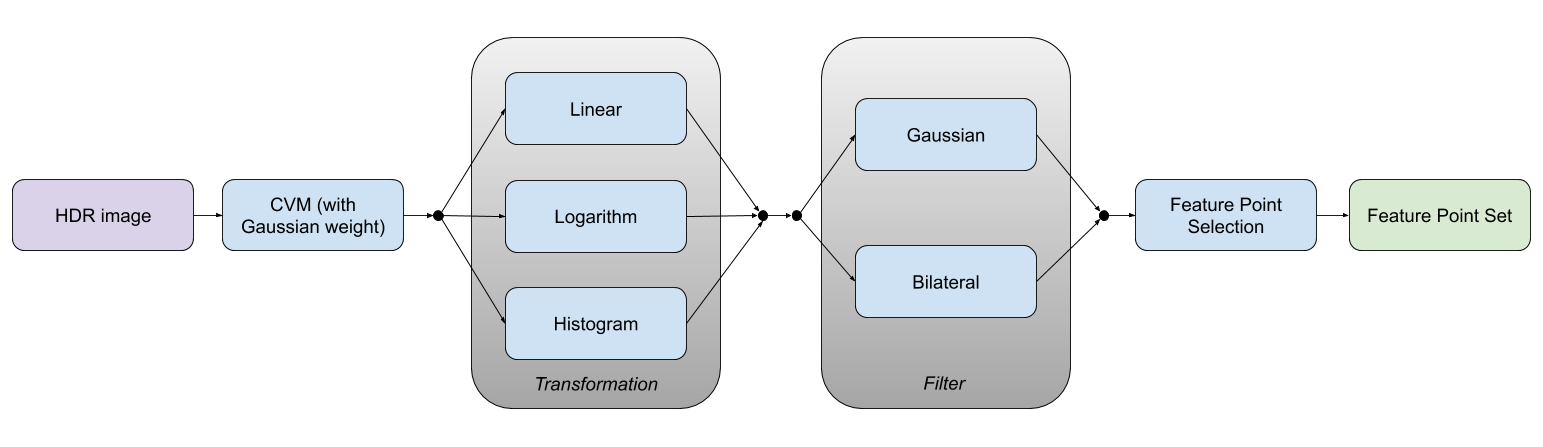}
    \caption{Selection test flow chart to choose techniques to compose DetectorCV algorithm.}
    \label{fig:selectionTest_flow}
\end{figure*}

DetectorCV algorithm has four main steps (Figure~\ref{fig:detectorCV_flow}), however, we have three \textit{transformations} and two \textit{filters} to choose, which gives us in total six possible combinations. To select the technique that maximizes our FP evaluation criteria, we tested all the possible combinations of techniques, as shown in Figure~\ref{fig:selectionTest_flow}. 

\textit{Transformation} options are linear transformation, logarithmic transformation, and histogram equalization.
\begin{itemize}
    \item \textit{Linear transformation}: single pixel-wise multiplication by a constant defined empirically as $25$. 
    \item \textit{Logarithmic transformation}: represented by $c \times \log(f (x, y))$, where $f(x, y)$ is the input image and $c$ is an constant set, also empirically, to $150$.
    \item \textit{Histogram equalization}: a well-known technique and does not have parameters. 
\end{itemize}

The empirical values mentioned above were defined by observing the results of a group of images from the datasets used.

\textit{Filter} options are Gaussian filter, and bilateral filter.

\begin{itemize}
    \item \textit{Gaussian filter}: we compared three filter mask sizes: $5\times5$, $9\times9$, and $15\times15$, based on image average size and in a previous study~\cite{pvribyl2016}. Based on the OpenCV\footnote{Open Source Computer Vision (OpenCV) Library. Documentation: https://docs.opencv.org/. Last access on December 2020.} implementation of this method, the value for each square mask is a function of its side (Equation~\ref{eq_k}).
    
\begin{equation}
\sigma_f = 0.3 ((\operatorname{side}-1)0.5 - 1) + 0.8
\label{eq_k}
\end{equation}

    \item \textit{Bilateral filter}: considering the elevated level of noise filtering we wanted to achieve, we tried the bilateral filter with mask size $10 \times 10$. To achieve a strong filter effect, three different values for the \textit{sigma space} parameter were compared: $150$, $175$, and $200$; \textit{sigma color} parameter was equal to sigma space parameter. 
\end{itemize}

Additionally, in the CVM step we tested four different options: one option without the Gaussian filter; and three values for $\sigma_c$ in the Gaussian filter (as seen in Equation~\ref{eq:2dgauss}): $1.0$, $1.5$, $2.0$. Considering this range of $\sigma_c$ values, the Gaussian mask size was $5\times5$ and the population size was $25$ ($5 \times 5$). These values were chosen to achieve the blurring effect we desired while maintaining a reduced mask size.

The detection algorithms classify the FPs by response strength, calculated in intermediate steps of the algorithm. Regarding the FP selection, we follow the same approach proposed by P{\v{r}}ibyl et al.~\cite{pvribyl2016}. We select the $500$ candidates with the strongest response and consider that the size of the squared region where non-local maximal suppression is applied is $21\times21$. The size of the squared region was chosen to approximate $0.5\%$ of the average size of the images from the databases.

We have then a total of $72$ possible combinations of parameters for the algorithm: 
\begin{itemize}
    \item \textbf{CVM}: four configurations (one without the Gaussian filter and three with different values for $\sigma_c$);
    \item \textbf{Transformation}: three options (Linear, Logarithmic and Histogram Equalization);
    \item \textbf{Filter}: six alternatives ($3$ configurations of the Gaussian filter and $3$ configurations of the Bilateral filter);
    \item  \textbf{Feature Point Selection}: single approach.
\end{itemize}   
    
These $72$ combinations were tested with each of the $43$ HDR images as well as with the $43$ logHDR images from both 2D Lighting and 3D Lighting datasets. We calculated uniformity and RR for each tested combination and generated 6192 tuples (uniformity, RR) as a result.  

The combination that optimizes both uniformity and RR criteria was chosen based on Pareto dominance. Considering the tuple (uniformity, RR) as coordinates of a vector, the vector Pareto dominance is formally defined as follows: vector $u=(u_1, u_2)$ is said to dominate a vector $v=(v_1, v_2)$ if and only if $u_1 > v_1$ and $u_2 > v_2$. More details are given by Coello~\cite{coello2007evolutionary}.

\section{Experimental Results and Discussion}

\subsection{DetectorCV steps optimization}

\begin{table}[b]
\begin{center}
\begin{tabular}
{|>{\centering\arraybackslash} m{2.0cm}|>{\centering\arraybackslash} m{1.8cm}|>{\centering\arraybackslash} m{0.6cm}|>{\centering\arraybackslash} m{1.1cm}|>{\centering\arraybackslash} m{1.2cm}|}
\hline
\textbf{Dataset-Type}                                       & \textbf{\backslashbox{Filter}{Transf.}}         & \textbf{\begin{tabular}[c]{@{}c@{}}Linear \\ \end{tabular}} & \textbf{\begin{tabular}[c]{@{}c@{}}Logarithmic \\ \end{tabular}} & \textbf{\begin{tabular}[c]{@{}c@{}}Histogram \\ Equalization\end{tabular}} \\ \hline
 
\multirow{2}{*}{3DLighting-HDR}      & Gaussian   & 151  & 64   & 83  \\ \cline{2-5} 
                                     & Bilateral & 66   & 100  & 168 \\ \hline

\multirow{2}{*}{3DLighting-logHDR}   & Gaussian  & 36   & 35   & 276 \\ \cline{2-5} 
                                     & Bilateral & 78   & 40   & 231 \\ \hline

\multirow{2}{*}{2DLighting-HDR}      & Gaussian  & 83   & 217  & 161 \\ \cline{2-5} 
                                     & Bilateral & 157  & 411  & 297 \\ \hline

\multirow{2}{*}{ 2DLighting-logHDR}  & Gaussian  & 151  & 196  & 183 \\ \cline{2-5} 
                                     & Bilateral & 37   & 35   & 7   \\ \hline
                                     
\multirow{2}{*}{\textbf{Summation}}  & Gaussian  & \textbf{421}  & \textbf{512}  & \textbf{703} \\ \cline{2-5} 
                                     & Bilateral & \textbf{338}  & \textbf{586}  & \textbf{703} \\ \hline
\end{tabular}
\end{center}
\caption{Number of dominated vectors by each algorithm.}
\label{tab:selectionTestResult}
\end{table}

\begin{figure}[ht]
    \centering
    \begin{subfigure}[]{1\linewidth}
        \includegraphics[width=\linewidth]{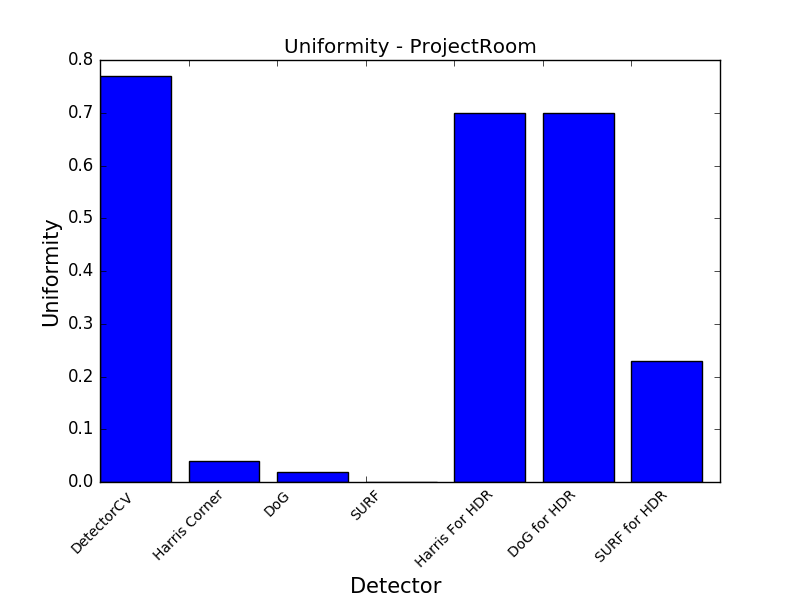}
        \caption{}
        \label{fig:result_uniformity}
    \end{subfigure}
    
    \begin{subfigure}[]{1\linewidth}
        \includegraphics[width=\linewidth]{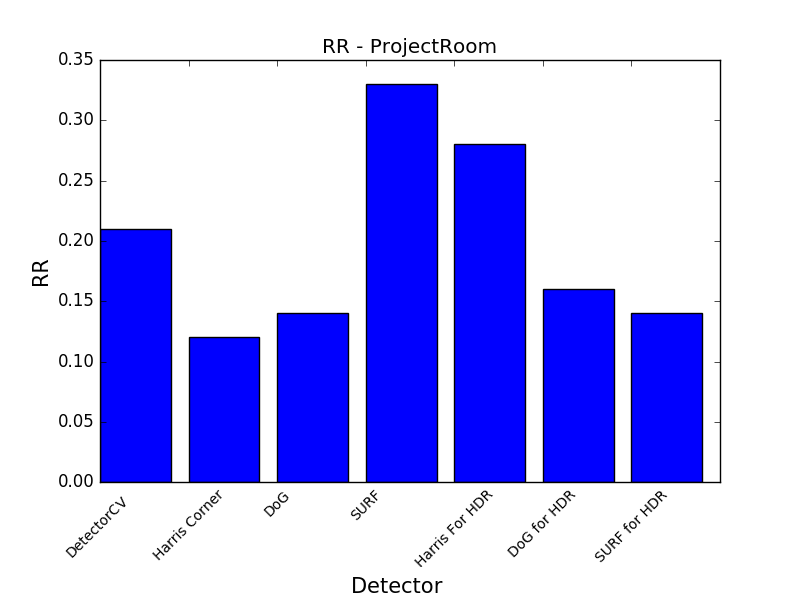}
        \caption{}
        \label{fig:result_RR}
    \end{subfigure}
    \caption{Experimental results: (a) uniformity; (b) repeatability rate. }
    \label{fig:results}
\end{figure}

To choose the best combination of steps to compose the DetectorCV algorithm, we obtained tuples with uniformity and RR for all possible combinations of dataset, \textit{transformation} and \textit{filter}. 
The vectors were grouped according to their input data. The input data is defined by a dataset-type pair, where the dataset is 2D Lighting and 3D Lighting, and the type is HDR or logHDR. We have then four groups: \textit{3D Lighting-HDR}, \textit{3D Lighting-logHDR}, \textit{2D Lighting-HDR}, and \textit{2D Lighting-logHDR}. Afterwards, we separated the vectors from each group in six subgroups based on the combinations of  techniques adopted during the implementation of the \textit{Transformation} and \textit{Filter} stages. The subgroups are: \textit{Linear+Gaussian}, \textit{Logarithmic+Gaussian}, \textit{Histogram Equalization+Gaussian}, \textit{Linear+Bilateral}, \textit{Logarithmic+Bilateral}, and \textit{Histogram Equalization+Bilateral}.

Inside each group, we counted how many vectors of different subgroups a vector dominates. The total for each subgroup is shown in Table~\ref{tab:selectionTestResult}. The last two rows of the table show the sum for each subgroup. 
Histogram Equalization has the best performance among \textit{Transformation} options. 
Summing all results from each filter, the Gaussian filter had a total of $1636$ dominated vectors; meanwhile, Bilateral had a total of $1627$ dominated vectors. So, we chose a Gaussian filter and histogram equalization to compose the algorithm.

The final body of the proposed DetectorCV algorithm, considering the best \textit{transformation} and \textit{filter} was: 

\begin{enumerate}
    \item \textbf{CVM}: with Gaussian weight ($\sigma_c = 2$, mask size = $5\times5$, population size = $25$); 
    \item \textbf{Transformation}: Histogram Equalization; 
    \item \textbf{Filter}: Gaussian (mask size = $9\times9$, $\sigma_f$ = 1.7); 
    \item \textbf{Feature Point Selection} (number of chosen candidates = 500; region of local maxima suppression = $21\times21$).
\end{enumerate}

To choose the Gaussian weight at the CVM step, for each dataset-type combination, we separated the vectors into four subgroups according to the four possible configurations for the CVM stage. For each subgroup, we calculated the total number of vectors that each vector in this subgroup dominates. The subgroup which dominates more vectors defines the winning combination for the Gaussian weight in the CVM step. The Gaussian filter size of 9$\times$9 was chosen as it dominates the groups with other mask sizes. The standard deviation value of the Gaussian filter chosen is derived from the mask size (Equation~\ref{eq_k}).


\subsection{Comparison to other algorithms}

The RR and uniformity measures of the best version of the DetectorCV algorithm were calculated in ProjectRoom dataset HDR images (Subsection~\ref{sec:dataset}). Results were compared to six other popular detectors: Harris Corner, DoG, SURF, Harris for HDR, DoG for HDR, and SURF for HDR. It is relevant to mention that this dataset was not used during the experiments performed to find the best implementation for DetectorCV, thus providing a neutral input. 

\textbf{Uniformity results}. DetectorCV has reached 0.77 of the maximum 1.0. In the  standard algorithms Harris Corner, DoG and SURF, the values obtained were 0.04, 0.02, and 0.00, respectively. Low uniformity, i.e., poor distribution of FPs among dark and bright areas of an image, using standard algorithms was expected. On the other hand, the modified algorithms Harris for HDR, DoG for HDR and SURF for HDR, proposed by Melo et al.~\cite{deMelo2018improving}, have reached 0.70, 0.70, and 0.23, respectively. These results are shown in Figure~\ref{fig:result_uniformity}. The results obtained with DetectorCV surpasses all the other algorithms, being at a level similar to the detectors modified for HDR (Harris Corner for HDR, and DoG for HDR), and almost 20 times higher than standard algorithms (Harris Corner, DoG, and SURF). 

\textbf{Repeatability Rate results}. DetectorCV algorithm has reached 0.21 of the maximum 1.0. In the  standard algorithms Harris Corner, DoG and SURF, the values obtained were 0.12, 0.14, and 0.33, respectively. The modified algorithms Harris for HDR, DoG for HDR, and SURF for HDR obtained 0.28, 0.16, and 0.14, respectively. These results are shown in Figure \ref{fig:result_RR}. We can observe that DetectorCV surpasses two of three of the standard algorithms, Harris Corner and DoG, as well as two of three of modified algorithms, DoG for HDR and SURF for HDR. Standard SURF algorithm have reached the maximum RR among all algorithms, followed by Harris for HDR.

\begin{figure}[ht]
    \centering
    \includegraphics[width=\linewidth]{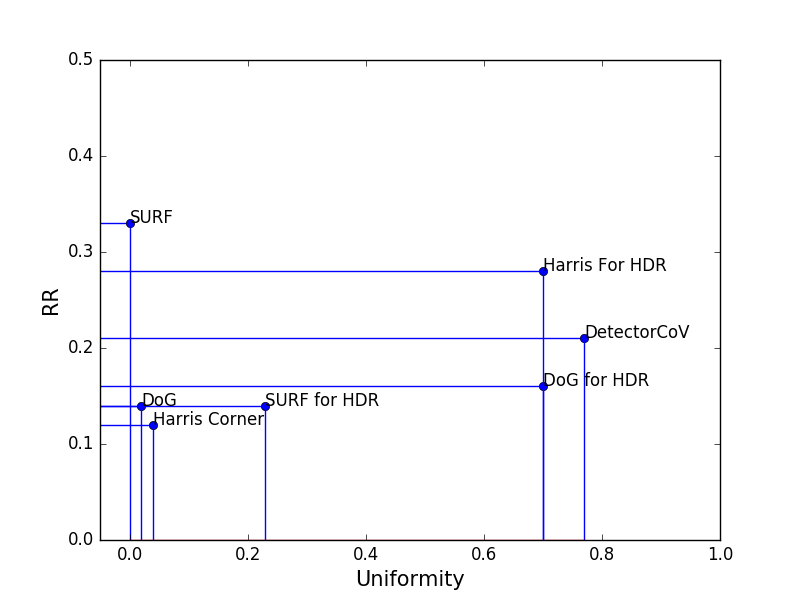}
    \caption{Vector dominance relationship.}
    \label{fig:result_dominance}
\end{figure}

\textbf{Vector dominance results}. Since we desire to maximize both the criteria, RR and uniformity, we need to analyze the results of RR and uniformity together. A way to analyze uniformity and RR together is through a vector dominance relationship. Figure~\ref{fig:result_dominance} shows uniformity and RR results as a vector for each algorithm applied to the ProjectRoom dataset. From the dominance definition, in the maximization case, we can observe which points are being dominated by another point. We can see that the nondominated vectors are the ones that represent SURF, Harris for HDR and DetectorCV algorithms. That way, these three algorithms are in Pareto front set, representing the optimal trade-off among all algorithms.

\begin{figure}[t]
    \centering
    \begin{subfigure}[t]{\linewidth}
        \includegraphics[width=\linewidth]{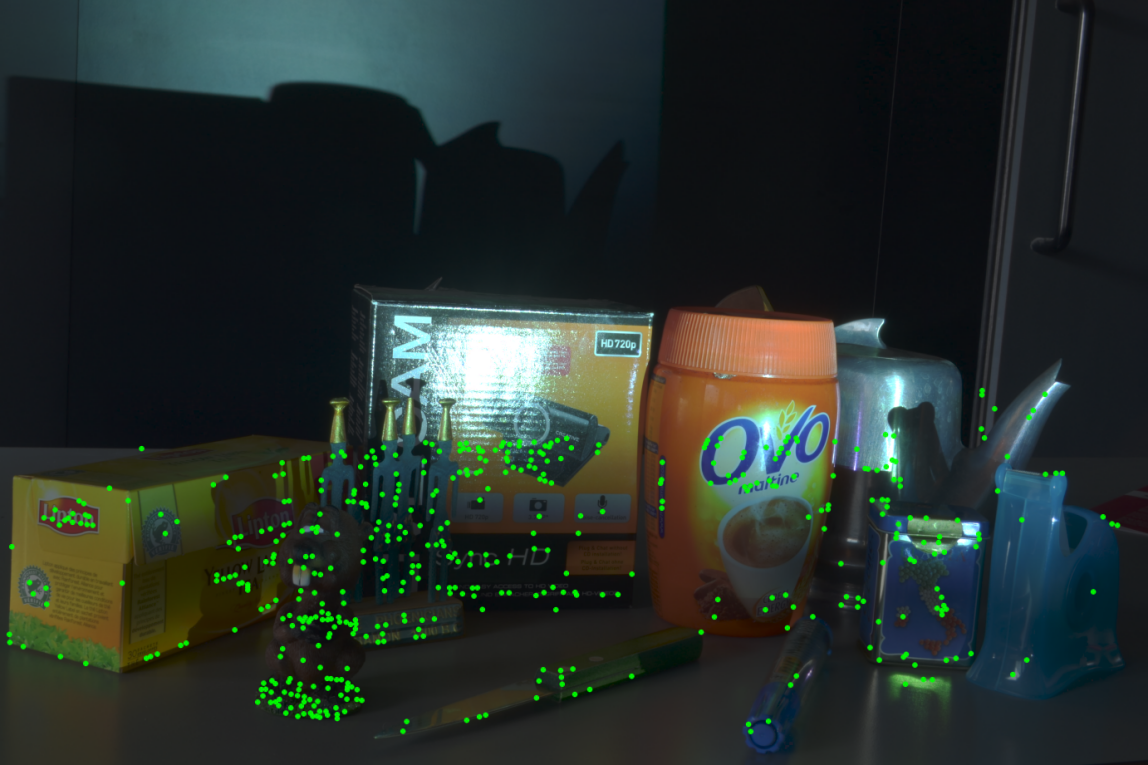}
        \caption{}
        \label{fig:ex_DetectorCV_FP}
    \end{subfigure}
    
    \begin{subfigure}[t]{\linewidth}
        \includegraphics[width=\linewidth]{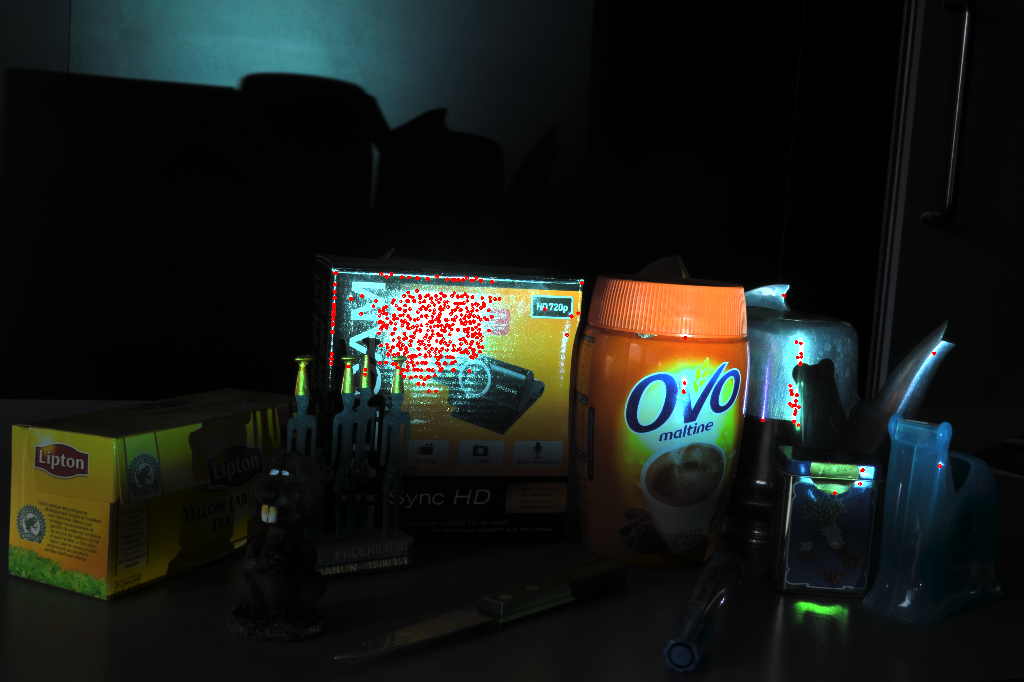}
        \caption{}
        \label{fig:ex_SURF_FP}
    \end{subfigure}
    \caption{Example of FP detection by (a) DetectorCV; and (b) SURF.}
    \label{fig:examplesFP}
\end{figure}

Analyzing the plot of dominance vectors, we can observe that, despite a high RR, the SURF algorithm has reached the lowest uniformity. This fact shows us that, despite most of the FPs can be found in a different image, most of these FPs are concentrated in highly illuminated areas. It means that SURF does not give us a good result, even being a nondominated vector. We can also see that SURF does not dominate any other algorithm, which is not the case of the other two nondominated vectors. Figure~\ref{fig:ex_DetectorCV_FP} shows an example of FPs found by DetectorCV. FPs are represented as green dots. At first glance, we can notice the high density of FPs found in dark areas of the image. Figure \ref{fig:ex_SURF_FP} shows us the detection result of the SURF algorithm in the same scene and the same number of FPs, represented as red dots, was detected, most of them in the bright area. The colors we adopted to illustrate the FPs are chosen to highlight their apperance in each image.

Harris For HDR, on the other hand, provided better results in both criteria. Given the results, we conclude that compared to SIFT and SURF, the Harris Corner algorithm provided better results because the dataset used has no affine or scale changes, just illumination changes. SIFT and SURF were created to deal with these types of changes, which are not handled by the Harris Corner algorithm. This fact usually makes the resulting RR of Harris worse than SIFT and SURF.

DetectorCV had results similar to Harris For HDR. We can see that the DetectorCV vector dominated the same number of vectors as the Harris Corner For HDR. Being one of the nondominated vector means that our proposed algorithm delivered remarkable results.

\section{Conclusions}

In this paper, we propose an FP detection algorithm for HDR images called DetectorCV. The algorithm, designed from scratch, aims to improve FP detection in HDR images exploring HDR characteristics. To this end, we applied the coefficient of variation mask (CVM) as the fundamental step in the DetectorCV pipeline. We optimized uniformity and RR of a filter and a histogram transformation for better adapting the CVM to the detection in HDR images. We compared DetectorCV with three standard detection algorithms found in literature, and three algorithms modified to improve FP detection in HDR images. We used datasets with HDR images in extreme light conditions and two FP evaluation criteria to demonstrate the performance of the DetectorCV. 

Based on the vector dominance results, the DetectorCV provided better FP detection results in HDR images compared to the standard algorithms DoG (SIFT detector), Harris Corner detector, and SURF detector. Compared to modified detectors, the DetectorCV provided better results than DoG for HDR, and SURF for HDR. We conclude that detecting FPs directly in HDR images is possible and can achieve excellent results as we better understand the behavior of pixel values in HDR images. 
In future studies, we intend to improve DetectorCV to detect FPs on different scales.


\section{Acknowloedgments}

This study was partially financed by Coordenação de Pesquisa da Universidade Federal de Sergipe (COPES-UFS) through the project PVB6533-2018.

\section*{Conflict of interest}

The authors declare that they have no conflict of interest.

\section*{Data availability}
The code developed during the current study is available on GitHub~\footnote{GitHub repository: \url{https://github.com/welersonMelo/Research-on-Feature-Point-Detector-HDRI}.}.

\bibliographystyle{IEEEtran}
\bibliography{references}

\end{document}